\documentclass[final]{cvpr}

\usepackage{times}
\usepackage{epsfig}
\usepackage{graphicx}
\usepackage{amsmath}
\usepackage{amssymb}

\usepackage{amsmath}
\usepackage{graphicx}
\usepackage{lineno}
\usepackage{array}
\usepackage{longtable}
\usepackage[numbers]{natbib}
\usepackage[usenames,dvipsnames]{color}
\usepackage{amssymb}
\usepackage{textcomp}
\usepackage{url}
\usepackage{float}
\usepackage{multirow}
\usepackage{microtype}

\usepackage{times}
\usepackage{epsfig}
\usepackage{graphicx}
\usepackage{amsmath}
\usepackage{amssymb}

\usepackage{subfig}
\usepackage{gensymb} 
\usepackage{booktabs}
\usepackage{multirow}
\usepackage{caption}
\usepackage{nccmath}
\usepackage{graphicx}
\usepackage{enumerate}

\newcommand{\rednew}[1] {\textcolor[rgb]{0.678,0.282,0.239}{{#1}}}
\newcommand{\bluenew}[1] {\textcolor[rgb]{0.341,0.313,0.639}{{#1}}}
\newcommand{\greennew}[1] {\textcolor[rgb]{0.302,0.639,0.373}{{#1}}}

\def\Put(#1,#2)#3{\leavevmode\makebox(0,0){\put(#1,#2){#3}}}

\usepackage{amsthm}

\theoremstyle{definition}

\def\ie{{\it i.e.}}
\def\eg{{\it e.g.}}
\def\etc{{\it etc.}}
\def\etal{{\it et~al.}}

\usepackage[pagebackref=true,breaklinks=true,colorlinks,bookmarks=false]{hyperref}

\def\cvprPaperID{4439} 
\def\confYear{CVPR 2021}

\begin{document}\sloppy

\title{FESTA: Flow Estimation via Spatial-Temporal Attention for Scene Point Clouds}

\def\namespacing{25pt}
\def\emailspacing{3pt}
\author{Haiyan Wang$^{1,2*}$\hspace{\namespacing}Jiahao Pang$^1$\thanks{Authors contributed equally. Work done while Haiyan Wang was an intern at InterDigital. }\hspace{\namespacing}Muhammad A. Lodhi$^1$\hspace{\namespacing}Yingli Tian$^2$\hspace{\namespacing}Dong Tian$^1$\\
$^1$InterDigital\hspace{30pt}$^2$The City College of New York\\
{\tt\small hwang005@citymail.cuny.edu,\hspace{\emailspacing}jiahao.pang@interdigital.com}\\
{\tt\small muhammad.lodhi@interdigital.com,\hspace{\emailspacing}ytian@ccny.cuny.edu,\hspace{\emailspacing}dong.tian@interdigital.com}
}

\maketitle
\pagestyle{empty}
\thispagestyle{empty}

\begin{abstract}
%
\vspace{-5pt}
Scene flow depicts the dynamics of a 3D scene, which is critical for various applications such as autonomous driving, robot navigation, AR/VR, etc.
Conventionally, scene flow is estimated from dense/regular RGB video frames.
With the development of depth-sensing technologies, precise 3D measurements are available via point clouds which have sparked new research in 3D scene flow.
Nevertheless, it remains challenging to extract scene flow from point clouds due to the sparsity and irregularity in typical point cloud sampling patterns.
One major issue related to irregular sampling is identified as the randomness during point set abstraction/feature extraction---an elementary process in many flow estimation scenarios.
A novel Spatial Abstraction with Attention (SA$^2$) layer is accordingly proposed to alleviate the unstable abstraction problem.
Moreover, a Temporal Abstraction with Attention (TA$^2$) layer is proposed to rectify attention in temporal domain, leading to benefits with motions scaled in a larger range.
Extensive analysis and experiments verified the motivation and significant performance gains of our method, dubbed as Flow Estimation via Spatial-Temporal Attention (FESTA), when compared to several state-of-the-art benchmarks of scene flow estimation.

\end{abstract}
\vspace{-5pt}

\vspace{-5pt}
\section{Introduction}\label{sec:intro}

Our world is dynamic.
To promptly predict and respond to the ever changing surroundings, humans are able to perceive a moving scene and decipher the 3D motion of individual objects.
This capability to capture and infer from scene dynamics is also desirable for computer vision applications. 
For instance,  a self-driving car can maneuver its actions upon perceiving the motions in its surroundings~\cite{menze2015object}; whereas a robot can exploit the scene dynamics to facilitate its localization and mapping process~\cite{alcantarilla2012combining}.
Moreover, with advances in depth-sensing technologies, especially the LiDAR technologies \cite{geiger2012we}, point cloud data have become a common configuration in such applications.

To describe the motion of individual points in the 3D world, \emph{scene flow} extends 2D optical flow to the 3D vector field representing the 3D scene dynamics~\cite{vedula1999three}.
Hence, just like 2D optical flow needs to be estimated from video frames comprising images~\cite{wedel2008efficient,herbst2013rgb}, 3D scene flow needs to be inferred from point cloud data~\cite{guo2020deep}.
However, it is non-trivial to accurately estimate scene flow from point clouds.

\begin{figure}[t]
  \vspace{-9.3pt}
  \centering\scriptsize
  \subfloat[FPS]{\includegraphics[width=0.24\linewidth]{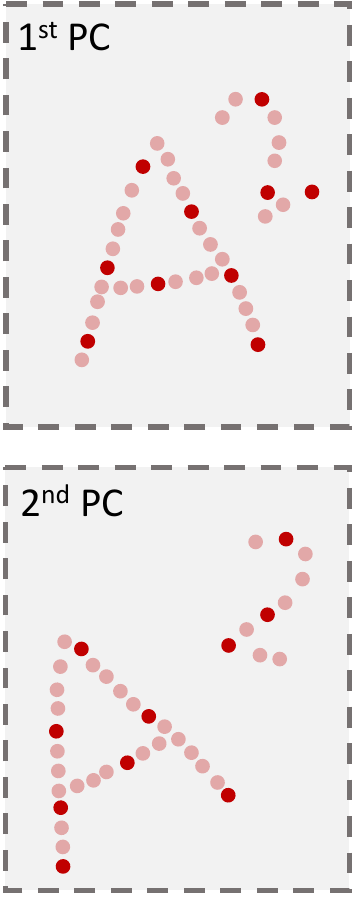}\label{fig:fps}}\hspace{1pt}
  \subfloat[Grouping]{\includegraphics[width=0.24\linewidth]{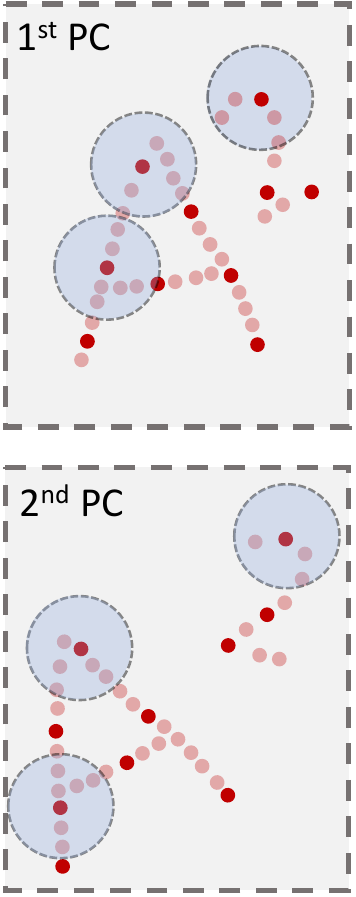}\label{fig:group_a}}\hspace{1pt}
  \subfloat[AP]{\includegraphics[width=0.24\linewidth]{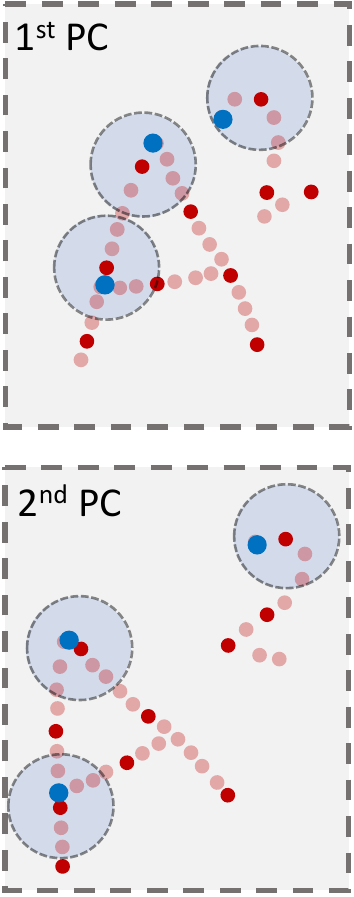}\label{fig:apstep}}\hspace{1pt}
  \subfloat[Regrouping]{\includegraphics[width=0.24\linewidth]{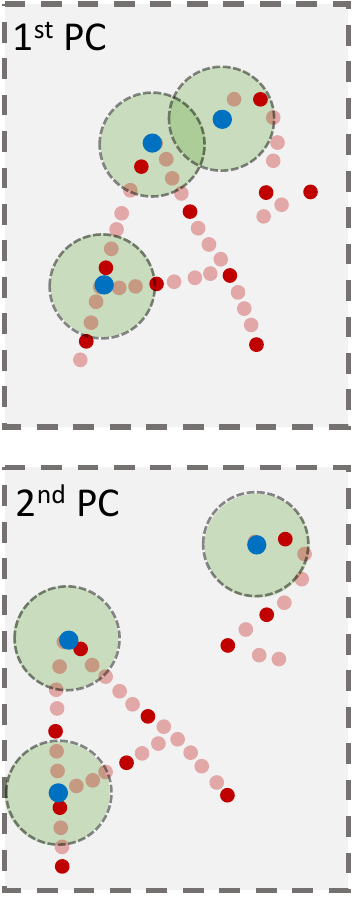}\label{fig:group_b}}
  \vspace{-5pt}
  \caption{\small Given two consecutive point clouds, down-sampled points produced by (a)~Farthest Point Sampling (FPS, in dark red) are different which make them intractable in scene flow estimation. However, by appending our (c)~Aggregate Pooling (AP), stable corresponding points (in blue) are synthesized for scene flow estimation.}
  \label{fig:spatial}
  \vspace{-15pt}
\end{figure}

\textbf{Unstable abstraction}:
Pioneered by PointNet \cite{qi2017pointnet} and its extension, PointNet++ \cite{qi2017pointnet++}, deep neural networks (DNNs) have recently been enabled to directly consume 3D point clouds for various vision tasks. 
As shown in Figure~\ref{fig:fps} and Figure~\ref{fig:group_a}, the grouping based on the Farthest Point Sampling (FPS) is widely utilized during the feature extraction process. It is treated as a basic point set abstraction unit for segmentation as well as scene flow estimation, \eg, FlowNet3D~\cite{liu2019flownet3d} and MeteorNet~\cite{liu2019meteornet}.
%
The naive FPS is simple and computationally affordable, but problematic.
Given two object point clouds, both representing the \emph{same} manifold, FPS would likely down-sample them differently~\cite{moenning2003fast} (see Figure~\ref{fig:fps}).
This inconsistency due to randomness in naive FPS is undesired for vision and machine tasks.
With two differently down-sampled point clouds, the subsequent grouping and abstraction would lead to two dissimilar sets of local features.
Thus, it becomes intractable to estimate the scene flow when comparing the features extracted via FPS.

To resolve this problem, we propose a \emph{Spatial Abstraction with Attention} (SA$^2$) layer which adaptively down-samples and abstracts the input point clouds. 
Compared to FPS, our SA$^2$ layer utilizes a trainable Aggregate Pooling (AP) module to generate much more \emph{stable} down-sampled points, \eg, blue points in Figure~\ref{fig:apstep}.
They define attended regions \cite{fu2017look} (\eg, green circles in Figure~\ref{fig:group_b}) for subsequent processing.

\textbf{Motion coverage}:
Similar to many deep matching algorithms for stereo matching and optical flow estimation, it is difficult to have a single DNN that can accurately estimate both large-scale motion and small-scale motion ~\cite{ilg2017flownet,pang2018zoom}.
To tackle this problem, we iterate the network for flow refinement with a proposed \emph{Temporal Abstraction with Attention} (TA$^2$) layer. It shifts the temporal attended regions to the more correspondent areas according to the initial scene flow obtained at the first iteration. 

In summary, we adaptively shift the attended regions when seeking abstraction from one point cloud spatially, and when fusing information across two point clouds temporally.
We name our proposal \emph{Flow Estimation via Spatial-Temporal Attention}, or FESTA for short. The main contributions of our work are listed as follows:
\vspace{-4pt}
\begin{enumerate}[(i)]
\item We propose the SA$^2$ layer for stable point cloud abstraction. It shifts the FPS down-sampled points to invariant positions for defining the attended regions, regardless of how the point clouds were sampled from the scene manifold. Effectiveness of the SA$^2$ layer is verified both theoretically and empirically.
\vspace{-4pt}
\item We propose the TA$^2$ layer to estimate both small- and large- scale motions. It emphasizes the regions that are more likely to find good matches between the point clouds, regardless of the scale of the motion.
\vspace{-4pt}
\item Our proposed FESTA architecture achieves the state-of-the-art performance for 3D point cloud scene flow estimation on both synthetic and real world benchmarks. Our method significantly outperforms the state-of-the-art methods of scene flow estimation.
\end{enumerate}
\section{Related Work}\label{sec:related}




Recent studies on scene flow estimation mainly extend methodologies for 2D optical flow estimation to 3D point clouds.
We first review the related research on optical flow estimation \cite{scharstein2002taxonomy}, then turn to deep learning methods for point cloud processing and scene flow estimation.

\textbf{Optical flow estimation}:
Optical flow estimation and its variant, stereo matching, both look for pixel-wise correspondence given a pair of 2D images.
Though conventionally solved with hand-crafted pipelines, recent proposals based on end-to-end DNNs achieve unprecedented performance.
Among these methods, FlowNet~\cite{dosovitskiy2015flownet,mayer2016large} is the very first trial, which adopts the popular hour-glass structure with skip connections.
This basic DNN architecture is remarkably successful for finding correspondence on images \cite{yang2019volumetric,chang2018pyramid}.
It is even extended to 3D point clouds for scene flow estimation, \eg, FlowNet3D~\cite{liu2019flownet3d} and HPLFlowNet\cite{gu2019hplflownet}.
However, it is difficult to estimate both small- and large-scale motions using one hour-glass architecture.
Thus a succeeding work, FlowNet2~\cite{ilg2017flownet}, stacks independent FlowNet modules to boost the performance, at the price of a larger model.
Differently, we resolve the problem with the TA$^2$ layer, which efficiently reuses part of the network for refinement.

\begin{figure*}
\begin{center}
\includegraphics[width=0.96\linewidth]{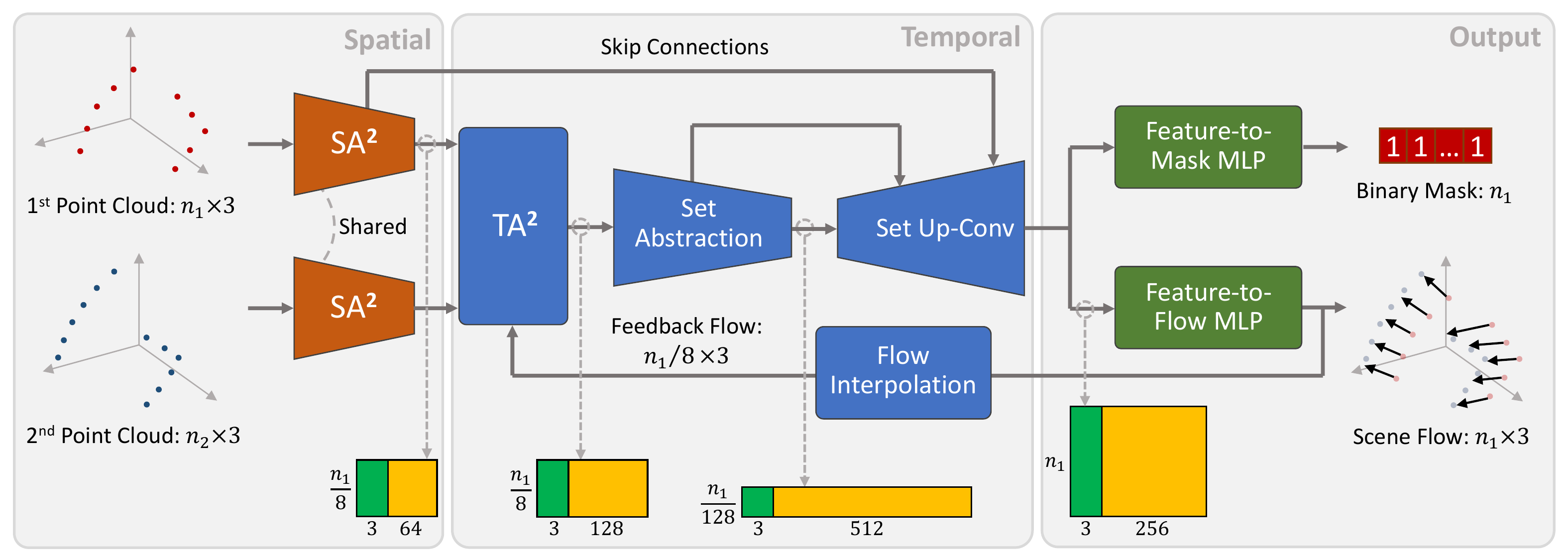}
\vspace{-5pt}
\caption{\small Our proposed FESTA architecture. On top of the FlowNet/FlowNet3D backbone, we specially incorporate the spatial-temporal mechanism with the proposed SA$^2$ and TA$^2$ layers.}
\label{fig:pipeline}
\end{center}
\vspace{-20pt}
\end{figure*}


\textbf{Deep learning on point clouds}:
%
Point cloud data is usually preprocessed, \eg, voxelized, so as to comply with deep learning frameworks justified for regular images/videos. 
Emerging techniques for native learning on point clouds relieve this need for format conversion.
The seminal work, PointNet~\cite{qi2017pointnet} directly operates on input points and produces a feature depicting the object geometry.
The learned features achieve point permutation-invariance through a pooling operation.
PointNet++~\cite{wang2020flownet3d++} applies FPS, followed by nearest-neighbor (NN) grouping and PointNet to abstract an input point cloud.
This abstraction step has become a popular elementary unit to digest point clouds.
Recent works, such as \cite{yang2019modeling,nezhadarya2020adaptive,lang2020samplenet}, propose complicated DNN architectures for the abstraction step; while our SA$^2$ layer is a lightweight module to serve the same purpose.
Moreover, these works limit themselves by \emph{selecting} existing points from the point cloud while we \emph{synthesize} new points to better represent the underlying geometry.


\textbf{Scene flow estimation}:
The task of 3D scene flow estimation was first introduced by Vedula~\etal~\cite{vedula1999three}.
It is conventionally estimated from RGB-D videos \cite{wedel2008efficient} or stereo videos \cite{herbst2013rgb}.
Only with the advent of deep learning, has the 3D scene flow estimation problem directly over point cloud data been enabled \cite{guo2020deep}.

FlowNet3D~\cite{liu2019flownet3d} is the first work directly learning scene flow from 3D point cloud data.
It ``converts'' the FlowNet~\cite{dosovitskiy2015flownet} architecture from the 2D image domain (with the convolutional neural network) to point cloud data (with PointNet).
A follow-up work, FlowNet3D++~\cite{qi2017pointnet++}, improves the performance by explicitly supervising the flow vectors with both their magnitudes and orientations; while the recently proposed PointPWC-Net~\cite{wu2019pointpwc} estimates the scene flow in a coarse-to-fine manner by fusing the hierarchical point cloud features.
Other notable methods include HPLFlowNet~\cite{gu2019hplflownet} applying the concepts of permutohedral lattice~\cite{adams2010fast} to extract structural information and \cite{mittal2020just} which is a self-supervised approach.
However, most of the efforts apply FPS to down-sample the input point clouds and introduce the unstable abstraction problem as mentioned.
In contrast, we propose the SA$^2$ layer to retrieve invariant down-sampled points, which greatly benefit the subsequent matching process.

\section{Framework Overview}\label{sec:overview}

\subsection{Architecture Design}
\label{sec:Architecture}
%
The architecture of our proposal, FESTA, is shown in Figure~\ref{fig:pipeline}, which follows the backbone of FlowNet3D \cite{liu2019flownet3d} and FlowNet~\cite{dosovitskiy2015flownet} with an hour-glass structure.
Each feature produced by network layers consists of a representative point accompanied by a local descriptor, \eg, the 
Spatial Abstraction with Attention (SA$^2$) layer generates $n_1/8$ such features for the first point cloud.
Given two input point clouds, they are respectively consumed by shared SA$^2$ layers to extract two sets of features,
which we call the \emph{spatial features}.
Then a proposed Temporal Abstraction with Attention (TA$^2$) layer serves as a coupling module to fuse the spatial features with the first point cloud serving as the reference.
Its output is another set of features that we called the \emph{temporal feature}.
Different from the spatial features, the temporal features fuse information of both point clouds, from which the 3D scene flow can be extracted.
After that, several Set Abstraction layers and Set Up-Conv layers from FlowNet3D~\cite{liu2019flownet3d} are appended to digest the temporal features, which complete the hour-glass structure.
The outputs of the last Set Up-Conv layer are a set of \emph{point-wise features}, associated with each point in the first point cloud.
To extract point-wise scene flow, we simply apply shared MLP layers to convert each point-wise feature to a scene flow vector.

Inspired by \cite{ilg2018occlusions}, we also estimate a binary mask indicating the existence of the scene flow vector for each point in the first point cloud.
In practice, the scene flow vectors may not be available due to occlusion and motion out of the field of view, \etc\,
The indication of their existence may serve as side information to help subsequent tasks.
Similar to the computation of 3D scene flow, dedicated MLP layers are applied to convert the point-wise features to the existence mask.

To enhance the scene flow estimation accuracy, especially to tackle motion of all ranges, we partially re-iterate our network with a feedback connection.
Though it is possible to run TA$^2$ many iterations, we found that running it twice achieves a good trade-off between computational cost and estimation accuracy.
Note that similar to \cite{liu2019flownet3d}, our FESTA architecture can be easily adapted to take additional attributes (\eg, RGB colors) as inputs.
Please refer to the supplementary material for more details of our architecture.

\subsection{Loss Function Design}

To effectively train the proposed FESTA in an end-to-end fashion, we evaluate both iteration outputs against the ground truth flow.
For each iteration, we first compute an $\ell_2$ loss between the ground-truth scene flow and the estimated one~\cite{liu2019flownet3d}.
This $\ell_2$ loss is denoted by $L_{\rm F}^{(i)}$, with $i$ being the iteration index.
Then the existence mask estimation is cast as a point-wise binary classification problem~\cite{ilg2018occlusions}, and 
a cross-entropy loss can be calculated against the ground-truth existence mask, denoted by $L_{\rm M}^{(i)}$.
The loss of the $i$-th iteration is finally given by:
\begin{equation}\label{eq:loss_it}
L^{(i)} = \mu L_{\rm F}^{(i)} + (1-\mu) L_{\rm M}^{(i)}.
\end{equation}
%
Our total loss for end-to-end training aggregates losses of both iterations, \ie,
\begin{equation}\label{eq:loss_tot}
L_{\rm tot}=(1-\lambda)L^{(1)}+\lambda L^{(2)}.
\end{equation}
Note that in \eqref{eq:loss_it} and \eqref{eq:loss_tot}, the hyper-parameters $\mu,\lambda\in [0,1]$.
Empirically, we set $\mu=0.8$ and $\lambda=0.7$.

\section{Spatial-Temporal Attention}\label{sec:attention} 


\subsection{Spatial Abstraction with Attention}\label{ssec:spatial}


\textbf{Design of the SA$^2$ layer}:
Key steps of our proposed Spatial Abstraction with Attention (SA$^2$) layer are illustrated in Figure~\ref{fig:spatial}. 
Farthest Point Sampling (FPS) followed by the Nearest Neighbor (NN) grouping (Figure~\ref{fig:fps}, Figure~\ref{fig:group_a}) are inherited from PointNet++~\cite{qi2017pointnet++} to divide the point cloud into groups as initial steps.
However, as mentioned in Section~\ref{sec:intro}, pure FPS-based abstraction produces \emph{unstable} down-sampled points that would tamper the scene flow estimation.

Herein, the design of the abstraction is motivated to reflect the \emph{intrinsic geometry} of the manifold $\mathcal{M}$, that is invariant to the randomness in the sampling pattern.
In this work, we approach this goal with a proposed Aggregate Pooling (AP) module.
Following FPS grouping, a point set is down-sampled using a synthesized point (Figure~\ref{fig:apstep}).
Then each newly down-sampled point defines its own attended region via another NN grouping step, leading to a new grouping scheme more suitable for a subsequent point-wise task (Figure~\ref{fig:group_b}), \ie, scene flow estimation in our case.
Similar to PointNet++~\cite{qi2017pointnet++}, finally we feed the new groups of points to a shared PointNet to extract their local descriptors.
The descriptors and the associated down-sampled points constitute the output of the SA$^2$ layer, \ie, the spatial features.
For example, see the $\frac{n_1}{8}\times 67$ matrix produced by the SA$^2$ layer in Figure~\ref{fig:pipeline}.

\begin{figure}[t]
\begin{center}
\includegraphics[width=0.96\columnwidth]{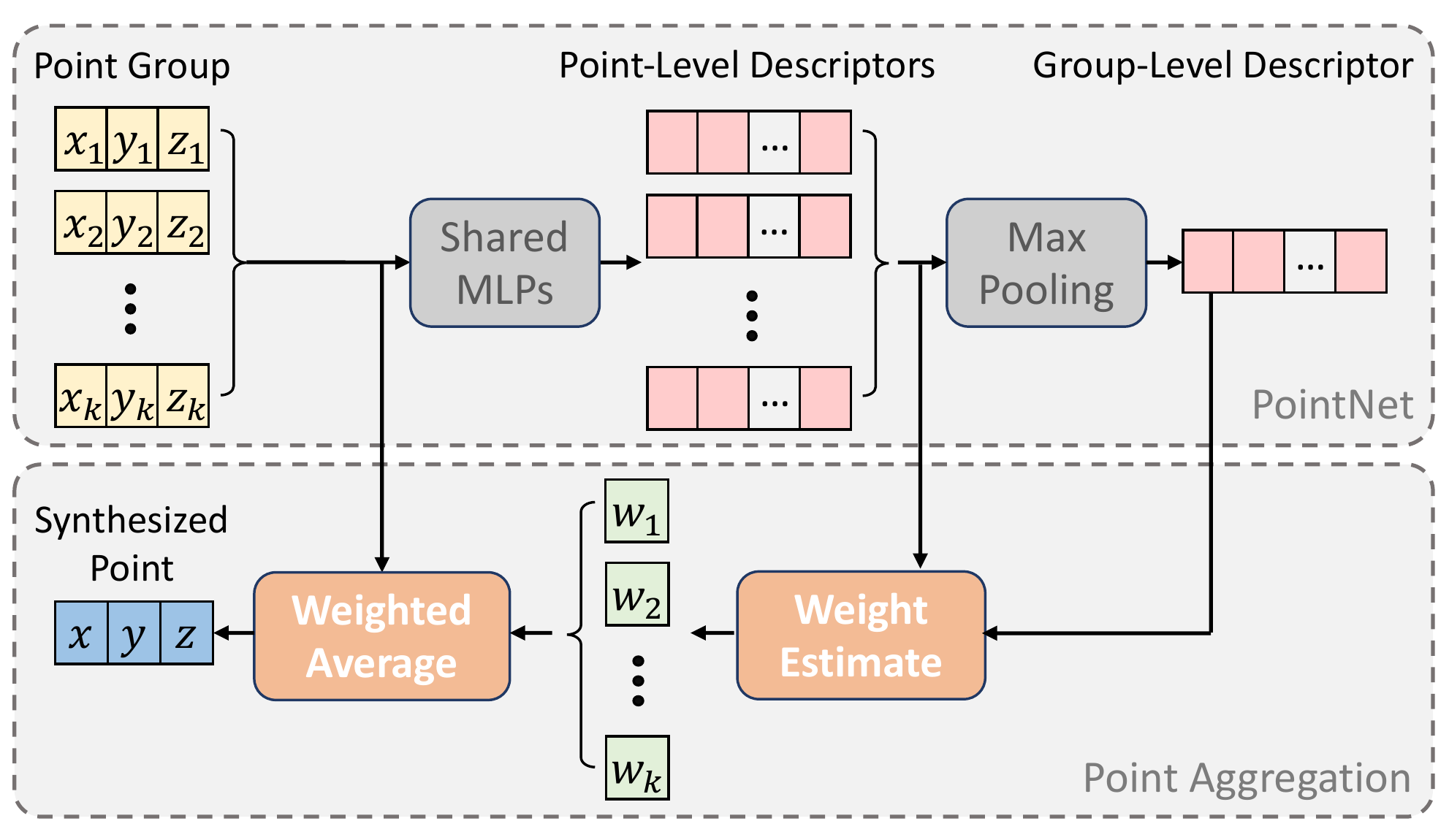}
\vspace{-5pt}
\caption{\small Block diagram of the Aggregate Pooling (AP) module.}
\label{fig:ap}
\end{center}
\vspace{-20pt}
\end{figure}

\textbf{Aggregate Pooling}:
The proposed AP module consumes a group of $k$ points and generates a synthesized point to represent the whole group.
As shown in Figure~\ref{fig:ap}, it consists of a PointNet~\cite{qi2017pointnet} and a point aggregation step.
The PointNet computes $k$ point-level descriptors with MLPs and a group-level descriptor with a max-pooling operator.
This PointNet is shared among all groups of points in the point cloud.
The point aggregation step then computes the weighted average of all the points in the group to synthesize a representative point.

Specifically, the point aggregation step measures the representativeness of a point (say, the $i$-th point) in the group with the similarity between its point-level descriptor (denoted by $\mathbf{f}_i$) and the group-level descriptor (denoted by $\mathbf{f}_{\rm g}$).
Among different ways to measure the vector similarity (\eg, Euclidean distance, correlation coefficient), we choose the dot-product metric like \cite{wang2018non} for its simplicity.
The obtained affinity values are then passed to a softmax function for normalization, resulting in a set of weights summed up to $1$.
Mathematically, the weight $w_i$ for the $i$-th point is
\begin{equation}\label{eq:weight}
w_i=\exp\left ( \mathbf{f}_i^\textrm{T}\mathbf{f}_\textrm{g} \right )\cdot \left [ \sum\nolimits_{j=1}^{k}\exp\left ( \mathbf{f}_j^\textrm{T}\mathbf{f}_\textrm{g} \right ) \right ]^{-1}.
\end{equation}
Suppose the points in the group are $\mathbf{s}_i=(x_i,y_i,z_i)$, $1\le i \le k$, then the synthesized point is simply $\sum\nolimits_{i=1}^{k}w_i\cdot\mathbf{s}_i$.

\textbf{Analysis}:
Now we attempt to understand the mechanism of the SA$^2$ layer to generate more stable points.
With the FPS and the grouping steps (Figure~\ref{fig:fps} and Figure~\ref{fig:group_a}), we have two lists of point sets from the input point cloud pair.
We first focus on a pair of point sets among the two lists that are sampled over the same Riemannian manifold patch $\cal M$ through the same sampling probability distribution $p(\mathbf{s}$).
Each point set is to be processed by the AP module.

Given sufficient number of points in both point sets characterizing the geometry $\cal M$, their group-level descriptors should be similar~\cite{qi2017pointnet}, denoted by $\mathbf{f}_g$.
In particular, if increasingly many 3D points are sampled from $\cal M$ according to $p$, then by definition, the synthesized points from both groups converge to the following integration on $\cal M$~\cite{kreyszig1978introductory}:
\vspace{-3pt}
\begin{equation}\label{eq:sync}
\mathbf{s}'=\frac{1}{\alpha}\int _{\cal M}w\left(\mathbf{f}(\mathbf{s})^\textrm{T}\mathbf{f}_g\right)p(\mathbf{s})\cdot\mathbf{s}\ d\mathbf{s},
\vspace{-3pt}
\end{equation}
where $\mathbf{f}(\mathbf{s})$ is the point-level descriptor of the point $\mathbf{s}$, function $w(\cdot)$ converts the dot-product measure $\mathbf{f}(\mathbf{s})^\textrm{T}\mathbf{f}_g$ to weight as is done by \eqref{eq:weight}, and $\alpha=\int _{\cal M}w\left(\mathbf{f}(\mathbf{s})^\textrm{T}\mathbf{f}_g\right)p(\mathbf{s})\ d\mathbf{s}$ is a normalization factor.
Please refer to the supplementary material for more detailed analysis.

Since the AP module converges to a \emph{fixed} location $\mathbf{s}'$ over $\cal M$, the SA$^2$ layer is expected to converge over a point cloud scene.
Empirical evidence is to be provided in Section~\ref{ssec:exp_abs} through a segmentation experiment.

Lastly, as the weight $w\left(\mathbf{f}(\mathbf{s})^\textrm{T}\mathbf{f}_g\right)$ is computed by a learnable network, the SA$^2$ layer is adaptable to down-stream tasks by generating novel, task-aware down-sampled points.


\subsection{Temporal Abstraction with Attention}\label{ssec:temporal}
\begin{figure}[t]
\begin{center}
\includegraphics[width=0.9\columnwidth]{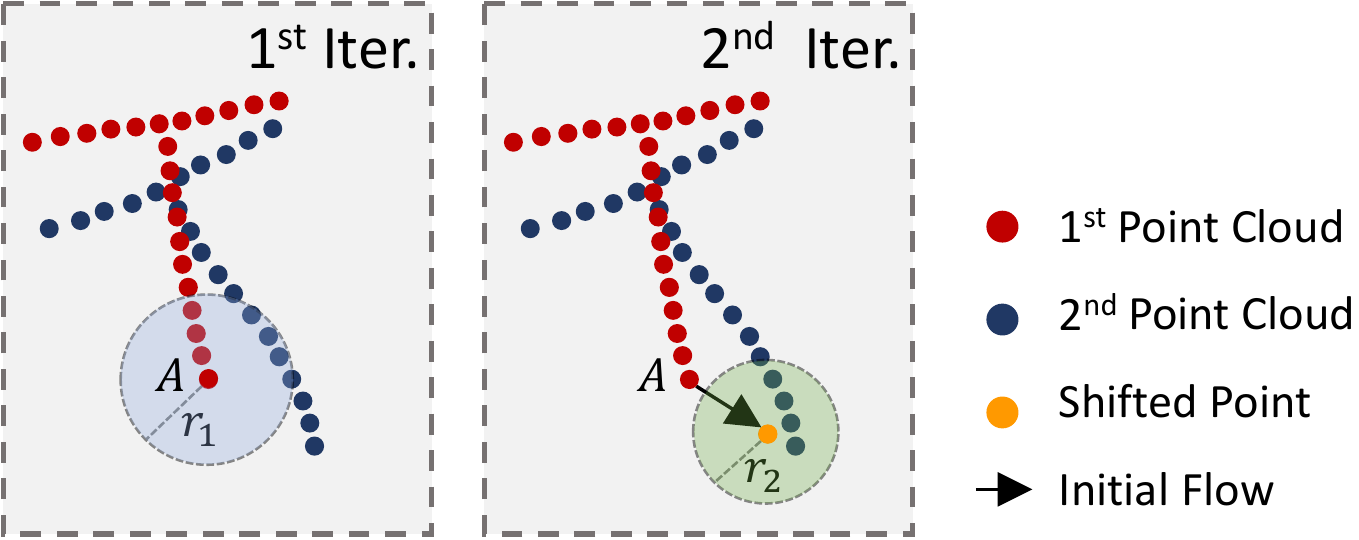}
\caption{\small Two iterations of the TA$^2$ layer. The blue circle in the left one demonstrates the attended region for the 1$^\textrm{st}$ iteration, which is inaccurate for the points correspondence. The green circle in the right one drags the attended region to the correspondent area by the initial flow estimated in the 1$^\textrm{st}$ iteration.}
\label{fig:temporal}
\end{center}
\vspace{-20pt}
\end{figure}

%
\textbf{Mechanisms of the TA$^2$ layer}:
The Temporal Abstraction with Attention (TA$^2$) layer aims to aggregate the spatial features of both point clouds given an initial scene flow.
During the first iteration without an initial scene flow presented, it behaves the same as the Flow Embedding layer in FlowNet3D \cite{liu2019flownet3d}.
Specifically, for each point (say, $A$) in the first down-sampled point cloud, we first perform a NN grouping step from the \emph{second} point cloud, which forms a group of points (centered at $A$) from the second point cloud, as shown in the left of Figure~\ref{fig:temporal}.
Then the grouped points, the point $A$ and its associated descriptor, are sent to a subsequent PointNet to extract another local descriptor.
More details of this extraction step can be found at \cite{liu2019flownet3d}.
The down-sampled point cloud and the new descriptors form the temporal features, \eg, the $\frac{n_1}{8}\times 131$ matrix produced by the TA$^2$ layer in Figure~\ref{fig:pipeline}.

During the second iteration, we reuse the spatial features generated by the SA$^2$ layers at the first iteration and feed them to the TA$^2$ layer (Figure~\ref{fig:pipeline}).
However, now with an initial scene flow corresponding to the first down-sampled point cloud available, we translate the search regions according to each of the scene flow vectors.
Specifically, suppose the coordinate of $A$ is $(x_A,y_A,z_A)$ and its initial scene flow vector is $(u_A, v_A, w_A)$, then the NN grouping is performed centered at $(x_A+u_A,y_A+v_A,z_A+w_A)$, see the right of Figure~\ref{fig:temporal}.
Note that to obtain the initial scene flow for the first down-sampled point cloud, it requires an extra interpolation step, \ie, the Flow Interpolation module in Figure\,\ref{fig:pipeline}.
We implement it as a simple deterministic module without trainable parameters.
To estimate the scene flow vector at a certain point, it computes the average scene flow vector in the neighborhood of that point.
Please refer to the supplementary material for more details.

\textbf{Analysis}:
Intuitively, the NN grouping step search for all points in the second point cloud that ``appear close'' to $A$; its search range defines an \emph{attended region} (the blue circle in Figure~\ref{fig:temporal}).
By only estimating the scene flow in one pass, it requires to choose a \emph{universal} search radius for all ranges of motion.
However, when the attended region (or equivalently, the search radius $r_1$ in Figure~\ref{fig:temporal}) is too small, it fails to capture large-scale motion; while for a large attended region or a large $r_1$, it includes too many candidates from the second point cloud and harms the granularity of the estimation (especially for small-scale motion).
This problem generally exists for not only scene flow estimation but also related problems such as stereo matching~\cite{pang2018zoom} and optical flow estimation~\cite{ilg2017flownet}. 

By introducing a second iteration, our TA$^2$ layer accordingly shifts the attended regions to confident areas that are more likely to observe good matches from the second point cloud.
Consequently, for the first iteration, it is more critical to identify a ``correct'' direction than a ``correct'' result. It is reflected in the selection of hyper-parameter $\lambda=0.7$.
Moreover, with a rough knowledge about how the second point cloud moves, the attended region at the second iteration (or radius $r_2$ in Figure~\ref{fig:temporal}) can be further reduced to search for more refined matching candidates.
\section{Experimentation}\label{sec:results}
This section first verifies how the SA$^2$ layer serves as a stable point cloud abstraction unit.
Then the proposed FESTA architecture is evaluated for 3D scene flow estimation.
Finally, we inspect how the key components contribute to the FESTA framework by an ablation study.

\subsection{Abstraction with the SA$^2$ Layer}\label{ssec:exp_abs}
The proposed SA$^2$ layer generally provides an alternative abstraction than FPS-like methods. 
It is desired to explicitly study its stability as pointed out in Section~\ref{ssec:spatial}.
In this test, we design a dedicated object segmentation process, as this per-point task has minimum additional procedure than abstraction, if compared to FESTA framework.
Note that the segmentation serves as a test bed to verify the stability of SA$^2$, rather than to claim state-of-the-art segmentation.



\textbf{Set-up}:
We build up a scene point cloud dataset using object point clouds from ModelNet40~\cite{wu20153d}.
This scene point cloud dataset contains 10$^4$ scenes; each scene contains $3$ to $6$ objects packed within a sphere-shaped container of radius $3$.
Moreover, all objects in a scene are normalized within a sphere of radius $r$, and are situated with a distance of at least $2$ between their object centers.
Evidently, by enlarging the objects with a bigger radius $r$, they are more likely to collide with one another, making it more challenging to distinguish and segment the objects.
We prepare $4$ versions of the scene point cloud dataset with different object radii $r$ ranging from $1$ to $1.8$.
Our object segmentation network is built upon the PointNet++~\cite{qi2017pointnet++} by replacing the FPS groupings with our proposed SA$^2$ layers.



\begin{figure}
\begin{center}
\includegraphics[width=1\linewidth]{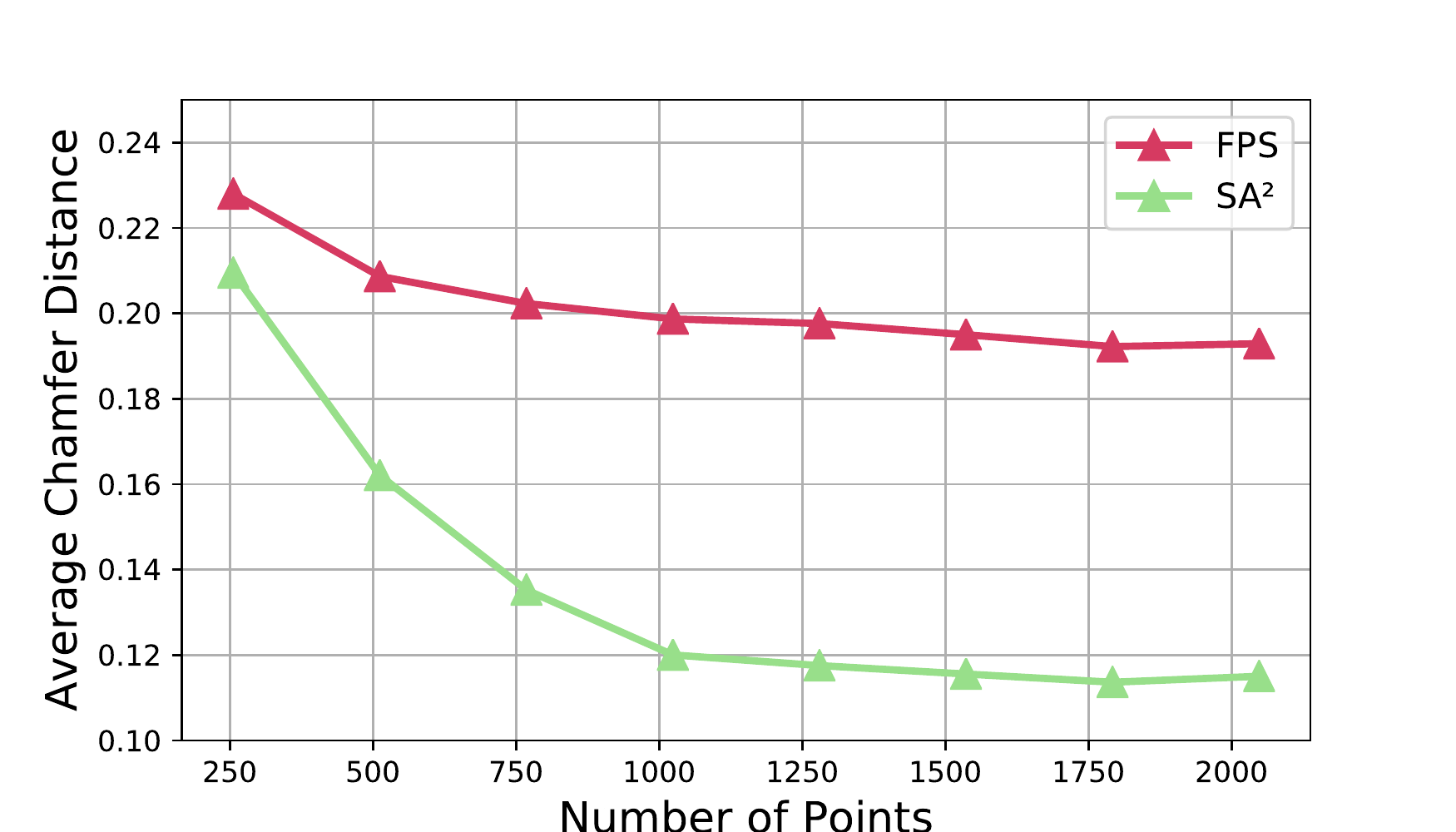}
\caption{\small Compared to FPS, our proposal generates more stable down-sampled point clouds.}
\label{fig:stable}
\end{center}
\vspace{-15pt}
\end{figure}

\textbf{Abstraction stability}:
We first evaluate the stability of our down-sampled point clouds using the above segmentation framework with dataset radius $r = 1.2$.
Given a scene in our multi-object dataset, we randomly pick $n\in[256, 2048]$ points from it for $100$ different times, resulting in $100$ input point clouds representing the same 3D scene. 
Then we feed these point clouds as inputs to both FPS-based and SA$^2$-based segmentation network to obtain down-sampled point sets containing only $64$ points.
For a stable down-sampling procedure, the down-sampled results from the $100$ point clouds should be similar to one another.
To evaluate the similarity, we compute the Chamfer Distances (CD)~\cite{fan2017point} between any two down-sampled point clouds, then take an average to characterize the stability.
A smaller average CD means more stable down-sampling.
Further averaging over $30$ scenes randomly selected from our dataset is performed.
Finally, as shown in Figure~\ref{fig:stable}, SA$^2$ always produces more stable down-sampled results than FPS.
Especially, for $n>1000$, our approach even reduces the average CD of FPS by about $50\%$, which confirms the superior stability of the proposed SA$^2$ layer.
The plot for SA$^2$ also verifies the analysis in Section~\ref{ssec:spatial}, which indicated that as the sampling becomes denser the down-sampled points become more stable.

\begin{figure}[t]
\begin{center}
\includegraphics[width=1\columnwidth]{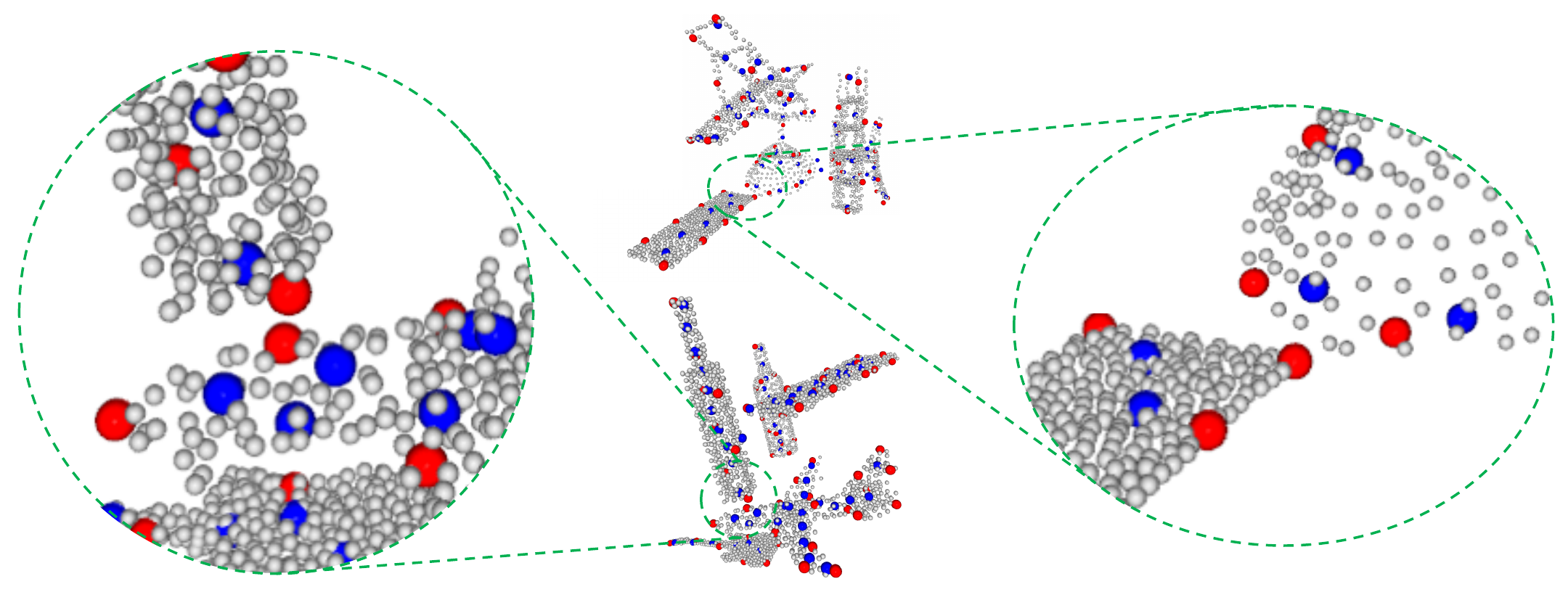}
\caption{\small Down-sampled points of FPS (red) and our SA$^2$ layer (blue) for the object segmentation task.}
\label{fig:segmentation}
\end{center}
\end{figure}

\begin{table}[t]
  \centering\footnotesize
  \caption{\small Object segmentation accuracy (\%).}\vspace{-8pt}
    \begin{tabular}{c||cccc}
    \hline
    \multirow{2}[4]{*}{\vspace{10pt}Methods} & \multicolumn{4}{c}{Object radius $r$} \\
\cline{2-5}          & 1.0   & 1.2   & 1.5   & 1.8 \\
    \hline
    \hline
    PointNet++ (FPS) & 92.56 & 88.74 & 65.87 & 43.07 \\
    Ours (SA$^2$) & \textbf{93.21}& \textbf{90.48} & \textbf{80.10} & \textbf{69.18} \\
    \hline
    \end{tabular}%
  \label{tab:mom}%
  \vspace{-5pt}
\end{table}%

\textbf{Evaluation}:
Having tested the SA$^2$ layer, we turn to understand how its stable abstraction benefits the segmentation performance.
Based on FPS and the SA$^2$ layer, two segmentation networks are trained on all the $4$ versions of the dataset (object radii $r$ ranging from $1$ to $1.8$) with the cross-entropy loss.
Table~\ref{tab:mom} compares the performance of SA$^2$ based segmentation with FPS-based segmentation, \ie, PointNet++, showing that SA$^2$ always provides higher segmentation accuracies.
As segmentation difficulty increases, our approach out-performs PointNet++ by a larger margin, \eg, our accuracy is 26\% higher than PointNet++ when $r=1.8$.

We demonstrate the down-sampled points of scene point clouds in Figure~\ref{fig:segmentation}, where the grey points depict the input point cloud; while the red and the blue points are those sampled by FPS and SA$^2$, respectively.
Thanks to the stable abstraction as verified earlier, the SA$^2$ layer consistently generates points belonging to distinctive objects and exhibits clearer separation between objects, which is highly preferred for segmentation.

\begin{figure*}[t]
\begin{center}
\includegraphics[width=1\linewidth]{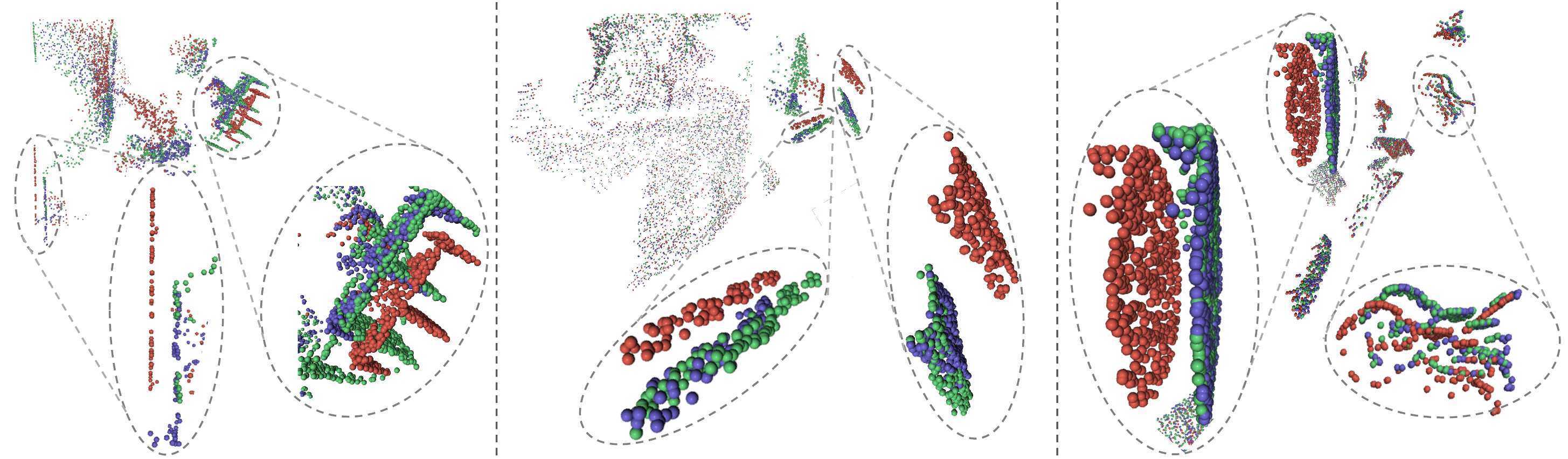}
\caption{\small Scene flow estimation on the FlyingThings3D dataset between 1$^\textrm{st}$ PC (in \rednew{red}), 2$^\textrm{nd}$ PC (in \greennew{green}). The results of our proposed FESTA architecture is shown with warped PC (in \bluenew{blue}) -- 1$^\textrm{st}$ PC warped by the scene flow.}
\vspace{-10pt}
\label{fig:f3d}
\end{center}
\end{figure*}

\begin{table*}[htbp]
  \centering\footnotesize
  \caption{\small Quantitative evaluation on the FlyingThings3D and the KITTI datasets.}\vspace{-2pt}
  \scalebox{1.0}{
    \begin{tabular}{c||ccc|ccc|ccc}
    \hline
    \multirow{2}[4]{*}{\vspace{10pt}Methods} & \multicolumn{3}{c|}{FlyingThings3D, geo.+RGB} & \multicolumn{3}{c|}{FlyingThings3D, geo.-only} & \multicolumn{3}{c}{KITTI, geo.-only} \\
\cline{2-10}          & EPE (m) & Acc S. ($\%$) & Acc R. ($\%$)& EPE (m)  & Acc S. ($\%$) & Acc R. ($\%$) & EPE (m)  & Acc S. ($\%$) & Acc R. ($\%$) \\
    \hline
    \hline
    FlowNet3D~\cite{liu2019flownet3d} & 0.1694 & 25.37 & 57.85 & 0.1705 & 23.71 & 56.05 & 0.1220 & 18.53 & 57.03 \\
    HPLFlowNet~\cite{gu2019hplflownet} & 0.1318     & 32.78     & 63.22     & 0.1453 & 29.46 & 61.91 & 0.1190 & 30.83 & 64.76 \\
    PointPWC-Net~\cite{wu2019pointpwc} & 0.1205     & 39.45     & 67.81     & 0.1310 & 34.22 & 65.78 & 0.1094 & 35.98 & 73.84 \\
    MeteorNet~\cite{liu2019meteornet} & -     & -     & -     & 0.2090 & -     & 52.12 & 0.2510 & -     & - \\
    FlowNet3D++~\cite{wang2020flownet3d++} & 0.1369 & 30.33 & 63.43 & 0.1553 & 28.50 & 60.39 & 0.2530 & -     & - \\
    Just~Go~w/~Flow~\cite{mittal2020just} & -     & -     & -     & -     & -     & -     & 0.1220 & 25.37 & 57.85 \\
    FESTA (Ours) & \textbf{0.1113} & \textbf{43.12} & \textbf{74.42} & \textbf{0.1253} & \textbf{39.52} & \textbf{71.24} & \textbf{0.0936} & \textbf{44.85} & \textbf{83.35} \\
    \hline
    \end{tabular}%
    }
  \label{tab:results}%
\end{table*}%


\subsection{Scene Flow Estimation with FESTA}\label{ssec:exp_sf}
\vspace{-0.5em}
Combining the SA$^2$ and the TA$^2$ layers, we evaluate the proposed FESTA architecture for scene flow estimation.

\textbf{Datasets}:
Our experiments are conducted on two popular datasets, the FlyingThings3D~\cite{mayer2016large} and the KITTI Scene Flow~\cite{geiger2012we} (referred to as KITTI) datasets.
Both of them are originally designed for matching tasks in the image domain (\eg, stereo matching).
Recently, Liu~\etal~\cite{liu2019flownet3d} converted them for scene flow estimation from 3D point clouds.
The FlyingThings3D dataset is a synthetic dataset with 20,000 and 2,000 point cloud pairs for training and testing, respectively.
In addition to the point cloud geometry, the RGB colors and the binary existence masks are also available.
Different from the FlyingThings3D, KITTI is a real dataset collected by LiDAR sensors, and contains incomplete objects.
The KITTI dataset has 150 point cloud pairs with available ground-truth scene flow.
Similar to \cite{liu2019flownet3d,gu2019hplflownet} and others, we only use the geometry (point coordinates) when computing the 3D scene flow.



\textbf{Benchmarks and evaluation metrics}:
We compare our FESTA with the following methods, listed in chronological order: FlowNet3D~\cite{liu2019flownet3d}, HPLFlowNet~\cite{gu2019hplflownet}, PointPWC-Net~\cite{wu2019pointpwc}, MeteorNet~\cite{liu2019flownet3d}, FlowNet3D++\cite{wang2020flownet3d++}, and a self-supervised method, Just~Go~with~the~Flow~\cite{mittal2020just}.
The scene flow quality is first evaluated with \emph{End-Point-Error} (EPE), which calculates the mean Euclidean distance between the ground-truth flow and the prediction. 
We also adopt two additional metrics from \cite{gu2019hplflownet}, \emph{Acc~Strict} and \emph{Acc~Relax}.\footnote{They are called Acc3D~Strict and Acc3D~Relax in \cite{gu2019hplflownet}. We remove ``3D'' since we do not need to distinguish with 2D cases.}
Both Acc~Strict and Acc~Relax aim to measure estimation accuracy, but with different thresholds.
Acc~Strict measures the percentage of points satisfying $\textrm{EPE} < 0.05m$ or $\textrm{relative error} < 5\%$; while Acc~Relax measures the percentage of points with $\textrm{EPE} < 0.1m$ or $\textrm{relative error} < 10\%$.

\textbf{Implementation details}:
The proposed FESTA is trained with two configurations over FlyingThings3D dataset following the FlowNet3D \cite{liu2019flownet3d}, with geometry-only and with additional RGB attributes. 
The two configurations are both trained with the Adam optimizer~\cite{kingma2014adam} for 500 epochs, with a batch size of 32 and a learning rate of 0.001. The size of the input point clouds are all set to 2048. 
All experiments are performed in the PyTorch~\cite{paszke2017automatic} framework.
For the geometry-only configuration, the inference is performed on both FlyingThings3D and KITTI.
In other words, the model is never tuned over KITTI, similarly done in \cite{liu2019flownet3d,gu2019hplflownet} and others.
With RGB attributes available, the inference is conducted on FlyingThing3D.
Quantitative and qualitative evaluations are performed.

\textbf{Quantitative evaluation}:
Quantitative results are reported in Table~\ref{tab:results}, where the proposed FESTA consistently outperforms the competing methods with significant gains.
For example, regarding to Acc Strict values for geometry only, our FESTA improves over the state-of-the-art method, PointPWC-Net, by 5.3\% for FlyingThings3D or 8.9\% for KITTI; and improves over our backbone, FlowNet3D, by 15.8\% for FlyingThings3D or 26.3\% for KITTI.
Encouragingly, when inspecting configurations with and without RGB over FlyingThings3D, it is noticed that in most cases, with geometry alone, our FESTA surpasses the competitors even when they take extra RGB attributes.
For instance, our FESTA (geometry only) achieves an EPE of 0.1253, lower than that of FlowNet++ (geometry+RGB) which is 0.1369.

We compare our model size and run time to representative methods and report in Table~\ref{tab:sizetime}, where the run time is evaluated on a Nvidia GTX 1080 Ti GPU with 11~GB memory.
We confirm that our superior performance is achieved by a model of size 16.1\,MB, similar to FlowNet3D which is 14.9\,MB.
It is much smaller than other competing methods, PointPWC-Net and HPLFlowNet.
Moreover, by removing the TA$^2$, we take similar run time as FlowNet3D, but still greatly reduce its EPE from 0.1705 (Table~\ref{tab:results}) to 0.1381 (to be seen in ablation study, Table~\ref{tab:ablation}).  

\begin{figure*}[t]
\begin{center}
\includegraphics[width=1\linewidth]{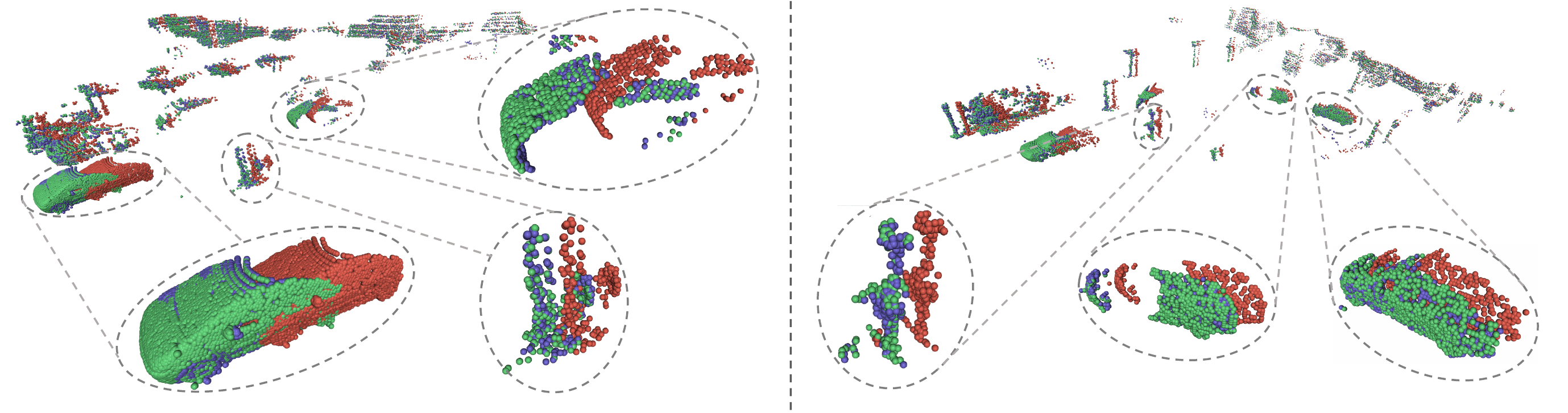}
\caption{\small Scene flow estimation on the KITTI dataset between 1$^\textrm{st}$ PC (in \rednew{red}), 2$^\textrm{nd}$ PC (in \greennew{green}). The results of our proposed FESTA architecture is shown with warped PC (in \bluenew{blue}) -- 1$^\textrm{st}$ PC warped by the scene flow.}
\vspace{-1.5em}
\label{fig:kitti}
\end{center}
\end{figure*}

\textbf{Qualitative evaluation}:
Visualization of the scene flow obtained by FESTA are shown in Figure~\ref{fig:f3d} for FlyingThings3D and in Figure~\ref{fig:kitti} for KITTI. Selected parts are zoomed in for a better illustration.
In each example, red points and green points represent the first and the second point cloud frames, respectively. Blue points represent the \emph{warped point cloud}, generated by translating each point in the first point cloud according to the estimated scene flow vector.
With more accurate scene flow vectors, the warped point cloud gets more overlapped with the second point cloud.
It can be seen that for all cases in Figure~\ref{fig:f3d}, \ref{fig:kitti}, our predicted scene flow produces warped point clouds that are highly overlapped with the second point cloud.
It affirms the effectiveness of our proposed FESTA on scene flow estimation.
Be reminded that our network has never observed any data from the KITTI dataset; however, it still successfully generalizes to KITTI and captures the dynamics solely based on the point coordinates.


%
%



\begin{table}[t]
  \centering\scriptsize
  \caption{\small Evaluation of model size and run time. F. - FlowNet3D \cite{liu2019flownet3d}; H. - HPLFlowNet \cite{gu2019hplflownet}; P. - PointPWC-Net \cite{wu2019pointpwc}.}\vspace{-5pt}
    \begin{tabular}{c|ccccc}
    \hline
    Metrics & F. & H. & P. & FESTA w/o TA$^2$ & FESTA (Ours)\\
    \hline
    Size (MB) & 14.9 & 231.8 & 30.1 & 16.1 & 16.1\\
    Time (ms) & 34.9 & 93.1 & 38.5 & 35.2 & 67.8\\
    \hline
    \end{tabular}%
  \label{tab:sizetime}%
  \vspace{-10pt}
\end{table}%

\subsection{Ablation Study}\label{ssec:ablation}
\vspace{-0.5em}
The ablation study is performed with geometry only configuration. We investigate the benefits of individual components in our FESTA architecture.
Specifically, we consider the following three variants:
\begin{enumerate}[(i)]
\vspace{-0.5em}
\item Replace the SA$^2$ layers as the simple FPS grouping followed by feature extraction in PointNet++~\cite{qi2017pointnet++} (or FlowNet3D~\cite{liu2019flownet3d});
\vspace{-0.5em}
\item Replace the TA$^2$ layer as the Flow Embedding layer in FlowNet3D, \ie, the second iteration is removed; and
\vspace{-0.5em}
\item Remove the outputting of existence mask, \ie, the network is trained only over 3D scene flow.
\vspace{-0.5em}
\end{enumerate}

Performance of the mentioned variants of FESTA are reported in Table~\ref{tab:ablation}.
The best- and the worst- performing metrics are highlighted with bold faces and with underlines, respectively.
We see that both the SA$^2$ and the TA$^2$ layers bring substantial gain to our model; while generally, the SA$^2$ layer is slightly more effective than the TA$^2$ layer.
Moreover, by jointly estimating an additional existence mask, our scene flow quality further improves.
That is because the ground-truth existence mask provides extra clues about the motion~\cite{ilg2018occlusions}, which supervises the network to capture the dynamics more precisely.

\begin{table}[t]
  \centering\footnotesize
  \caption{\small Evaluation of different variants of FESTA.}\vspace{-5pt}
    \begin{tabular}{c||ccc|ccc}
    \hline
    Data. & SA$^2$\hspace{-5pt} & TA$^2$\hspace{-5pt} & Mask  & EPE (m) & Acc S. ($\%$) & Acc R. ($\%$) \\
    \hline
    \hline
    \multirow{4}[2]{*}{Fly.} &    $\times$   & \checkmark & \checkmark & \underline{0.1402} & \underline{33.15} & \underline{66.14} \\
          & \checkmark &   $\times$    & \checkmark & 0.1381 & 34.70 & 67.36 \\
          & \checkmark & \checkmark &   $\times$    & 0.1289 & 37.91 & 69.05 \\
          & \checkmark & \checkmark & \checkmark & \textbf{0.1253} & \textbf{39.52} & \textbf{71.24} \\
    \hline
    \multirow{4}[2]{*}{KI.} &   $\times$    & \checkmark & \checkmark & \underline{0.1163} & \underline{30.10} & \underline{70.85} \\
          & \checkmark &   $\times$    & \checkmark & 0.1027 & 41.07 & 80.04 \\
          & \checkmark & \checkmark &   $\times$    & 0.0955 & 42.74 & 80.15 \\
          & \checkmark & \checkmark & \checkmark & \textbf{0.0936} & \textbf{44.85} & \textbf{83.35} \\
    \hline
    \end{tabular}%
  \label{tab:ablation}%
  \vspace{-8pt}
\end{table}%

\begin{figure}[t]
\begin{center}
\vspace{-14pt}
\includegraphics[width=1\linewidth]{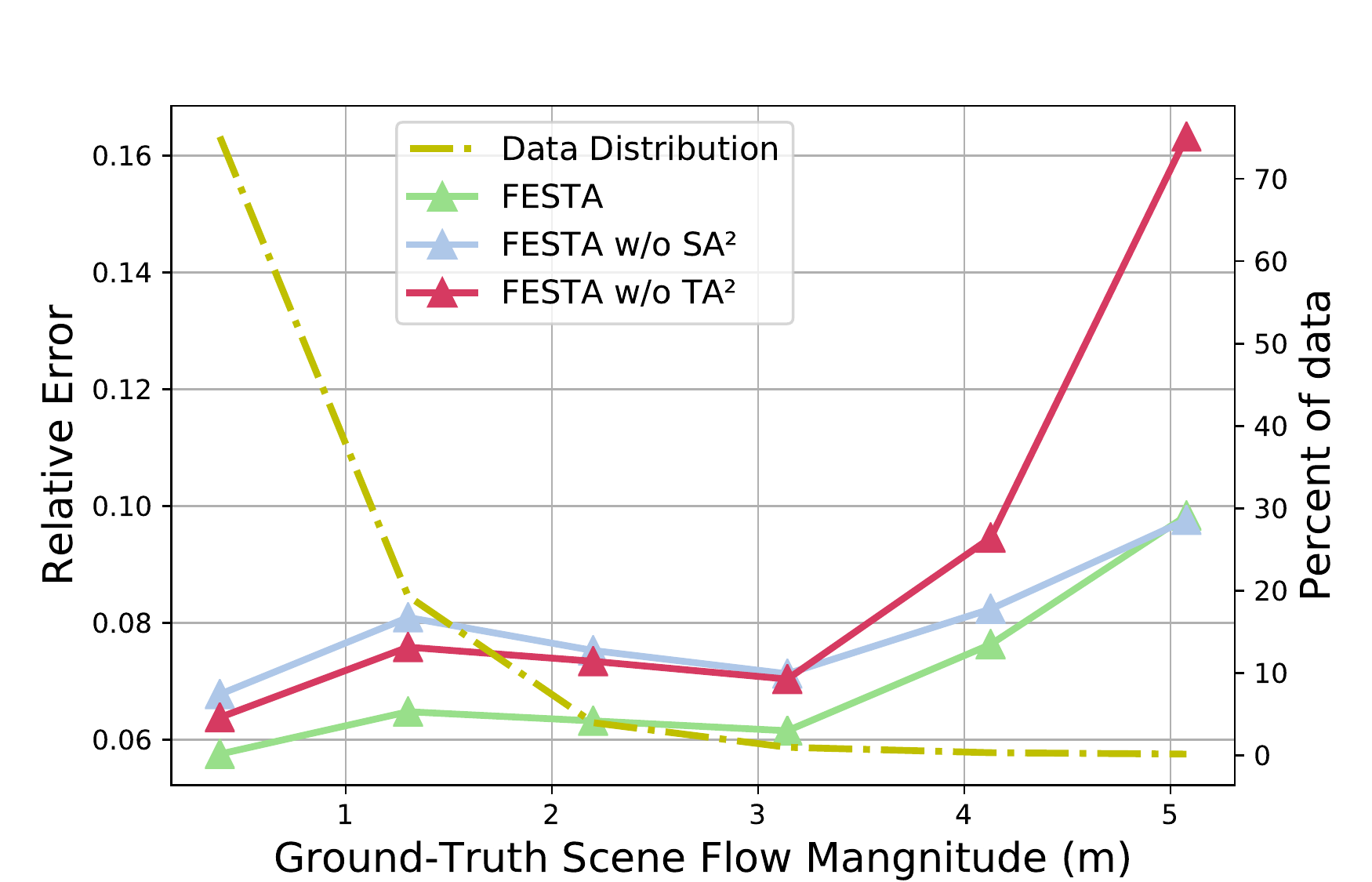}
\caption{\small For ground-truth motion with different magnitudes, the variants of FESTA perform differently.}
\vspace{-16pt}
\label{fig:ablation_range}
\end{center}
\end{figure}

Additionally, we further examine how the SA$^2$ and the TA$^2$ layers benefit the estimation of motion at different scales.
Specifically, we classify 3D points in the FlyingThings3D test set according to their ground-truth flow vector magnitudes.
For each bin of scene flow magnitude, we compute an average of relative error achieved by our FESTA, and count the bin size.
In this way, we plot a relative error curve (in green) of FESTA on different scene flow magnitudes, as shown in Figure~\ref{fig:ablation_range}.
We similarly plot the curves for the variants without SA$^2$ and TA$^2$ in blue and red, respectively.

By comparing the red and the green curves, we see that the TA$^2$ layer greatly improves the performance of large-scale motion, which is expected because the TA$^2$ directly shifts its attended region according to an initial scene flow.
Differently, the SA$^2$ layer benefits scene flow with smaller magnitudes.
That is because the SA$^2$ layer is able to adjust its attended region gently---the convex hull of an input point group (Section~\ref{ssec:spatial}), which enhances the granularity of the estimation and benefits mainly small-scale motion.
\vspace{-5pt}




\section{Conclusion}\label{sec:conclusion}
\vspace{-0.5em}
We propose a new spatial-temporal attention mechanism to estimate 3D scene flow from point clouds. The effectiveness of 
our proposed Flow Estimation via Spatial-Temporal Attention (FESTA) has been proven by our state-of-the-art performance of extensive experiments.
Essentially, our spatial-temporal attention mechanism successfully rectifies the region of interest (RoI) based on the feedback from earlier trials.
Its rationale resembles existing literature utilizing the attention mechanism, such as \cite{fu2017look} for image recognition and \cite{yin2016abcnn} for sentence modeling.
For future research, we plan to investigate the potentials of the SA$^2$ and TA$^2$ layers for different point cloud processing tasks, such as classification, registration, and compression. 


{\small
\bibliographystyle{ieee_fullname}

}

\end{document}